\documentclass[journal=jacsat,manuscript=article]{achemso}

\usepackage[version=3]{mhchem} % Formula subscripts using \ce{}
\usepackage{pifont}
\usepackage{booktabs}
\usepackage{comment}
\usepackage{tabularx}
\usepackage{hyperref}
\usepackage{subcaption}
\usepackage{listings}
\usepackage{xcolor}
\author{Genmao Zhuang}
\affiliation{Department of Materials Science and Engineering, Carnegie Mellon University, 5000 Forbes Ave, Pittsburgh, PA 15213, USA}

\author{Amir Barati Farimani}
\email{barati@cmu.edu}
\affiliation{Department of  Mechanical Engineering, Carnegie Mellon University, 5000 Forbes Ave, Pittsburgh, PA 15213, USA}

\title[MKNA for Evidence-Grounded Materials Discovery]
  {From Natural Language to Materials Discovery: The Materials Knowledge Navigation Agent
}

\abbreviations{MKNA, RAG, CGCNN, M3GNet, MP, DFT}

\begin{document}


% \begin{tocentry}

% Some journals require a graphical entry for the Table of Contents.
% This should be laid out ``print ready'' so that the sizing of the

% Inside the \texttt{tocentry} environment, the font used is Helvetica
% 8\,pt, as required by \emph{Journal of the American Chemical

% The surrounding frame is 9\,cm by 3.5\,cm, which is the maximum
% permitted for  \emph{Journal of the American Chemical Society}
% graphical table of content entries. The box will not resize if the
% content is too big: instead it will overflow the edge of the box.

% This box and the associated title will always be printed on a
% separate page at the end of the document.

% \end{tocentry}

\begin{abstract}
% The discovery of novel materials is the cornerstone of technological innovation in energy, electronics, and aerospace. Conventional approaches—based on high-throughput experiments or first-principles simulations—are costly and time-consuming, leaving vast regions of chemical space unexplored. Although machine learning accelerates property prediction, most models are task-specific and lack generalization across diverse materials problems. Here we introduce the Materials Knowledge Navigation
% Agent(MKNA), which couples large language models (LLMs) with materials databases, machine-learning predictors, and density functional theory (DFT) tools, enabling automatic feature retrieval, code generation, property prediction, and stability validation. Applied to high-Debye-temperature materials and low-band-gap semiconductors, the agent rapidly identifies larger sets of stable candidates, including both reported compounds and previously unexplored possibilities. This orchestration of retrieval, computation, and self-repair
% transforms expert-driven workflows into autonomous, scalable exploration. This approach demonstrates how LLM-powered agents can accelerate the path from data to knowledge, paving the way for autonomous scientific discovery in materials science.
Accelerating the discovery of high-performance materials remains a central challenge across energy, electronics, and aerospace technologies, where traditional workflows depend heavily on expert intuition and computationally expensive simulations. Here we introduce the \textit{Materials Knowledge Navigation Agent} (MKNA), a language-driven system that translates natural-language scientific intent into executable actions for database retrieval, property prediction, structure generation, and stability evaluation. Beyond automating tool invocation, MKNA autonomously extracts quantitative thresholds and chemically meaningful design motifs from literature and database evidence, enabling data-grounded hypothesis formation. Applied to the search for high--Debye--temperature ceramics, the agent identifies a literature-supported screening criterion ($\Theta_D > 800$~K), rediscovers canonical ultra-stiff materials such as diamond, SiC, SiN, and BeO, and proposes thermodynamically stable, previously unreported Be--C--rich compounds that populate the sparsely explored 1500--1700~K regime. These results demonstrate that MKNA not only finds stable candidates but also reconstructs interpretable design heuristics, establishing a generalizable platform for autonomous, language-guided materials exploration.

\end{abstract}

\section{Introduction}

% --- Polished MIT CommLab–style Introduction ---
The discovery of new materials has repeatedly enabled technological revolutions across energy storage, information processing, and aerospace engineering. Breakthroughs such as lithium-ion batteries \cite{goodenough2013li}, advanced semiconductor materials \cite{sze2008semiconductor}, and high-temperature alloys \cite{pollock2016alloy} illustrate a broader principle: materials innovation is not incremental optimization but the foundation upon which disruptive technologies are built. As technological demands intensify, the ability to rapidly identify and design functional compounds has become increasingly critical.

In response to this need, the past two decades have seen the rise of high-throughput experimental and computational platforms. Combinatorial synthesis techniques have enabled the parallel fabrication of thousands of samples \cite{xiang1995combinatorial,takeuchi2005combinatorial}, while high-throughput DFT workflows now allow systematic exploration of vast chemical spaces \cite{curtarolo2003predicting,curtarolo2005accuracy}. The resulting data are consolidated in large-scale repositories—including the Materials Project \cite{jain2013commentary}, AFLOW \cite{curtarolo2012aflowlib}, OQMD \cite{saal2013materials}, and NOMAD \cite{draxl2019nomad}—which provide open access to structures, formation energies, and derived properties for hundreds of thousands of compounds. Meanwhile, machine-learning models such as CGCNN \cite{xie2018crystal}, MEGNet \cite{chen2019graph}, M3GNet \cite{chen2022universal}, ElemNet \cite{jha2018elemnet}, and ALIGNN \cite{choudhary2021atomistic} have further accelerated computational screening by enabling accurate surrogate predictions, with recent advances including orbital-aware graph models and self-supervised pretraining for crystalline property prediction \cite{Karamad2020OGCNN,Magar2022CrystalTwins}.More recently, structure-agnostic and transformer-based pretraining has improved data efficiency and transferability across materials domains, spanning crystals, metal--organic frameworks, and polymers \cite{Huang2024PretrainingStrategies,Huang2024HeatCapacityTransformers,Cao2023MOFormer,Xu2023TransPolymer}.

Despite this progress, the discovery pipeline remains fragmented. Databases are incomplete \cite{jain2013commentary,draxl2019nomad}; machine-learning models are typically task-specific \cite{butler2018machine,schmidt2019recent,Huang2024PretrainingStrategies,Magar2022CrystalTwins}; workflows require substantial human intervention \cite{xie2019graph,jha2019enhancing}; and predictive modeling often fails to translate into experimental validation \cite{agrawal2016perspective,kalidindi2015materials}. Addressing these limitations requires a unified framework capable of reasoning across retrieval, prediction, and validation while autonomously extracting actionable scientific knowledge.

Recent work has begun exploring agent-based automation in materials science. Systems such as LLMatDesign \cite{LLMatDesign2024} generate hypotheses from natural-language prompts, HoneyComb \cite{HoneyComb2024} automates modular tool invocation, and other efforts focus on literature extraction \cite{Ansari2024} or autonomous laboratories \cite{szymanski2023autonomous}. However, these systems typically operate on isolated stages of the discovery pipeline and lack the ability to infer and utilize physicochemical priors to guide search.

To address these gaps, we introduce the \textit{Materials Knowledge Navigation Agent} (MKNA)—an autonomous scientific decision-maker that unifies semantic interpretation, knowledge retrieval, surrogate property estimation, structure generation, and physics-informed stability validation into a coherent closed loop. MKNA dynamically inspects available data, generates or repairs missing computations, and selectively invokes models such as CGCNN and M3GNet when needed, enabling robust progress even with incomplete databases.

We demonstrate MKNA's reasoning capability through a representative search objective: high--Debye--temperature materials. Starting from high-level natural-language queries, MKNA not only rediscovers known high-performance systems but also uncovers previously unreported candidates by inferring physicochemical priors (e.g., $\Theta_D>800$~K) and prioritizing lightweight covalent Be--C frameworks—revealing latent design rules rather than performing blind screening.

\begin{figure}[htbp]
    \centering
    \includegraphics[width=\textwidth]{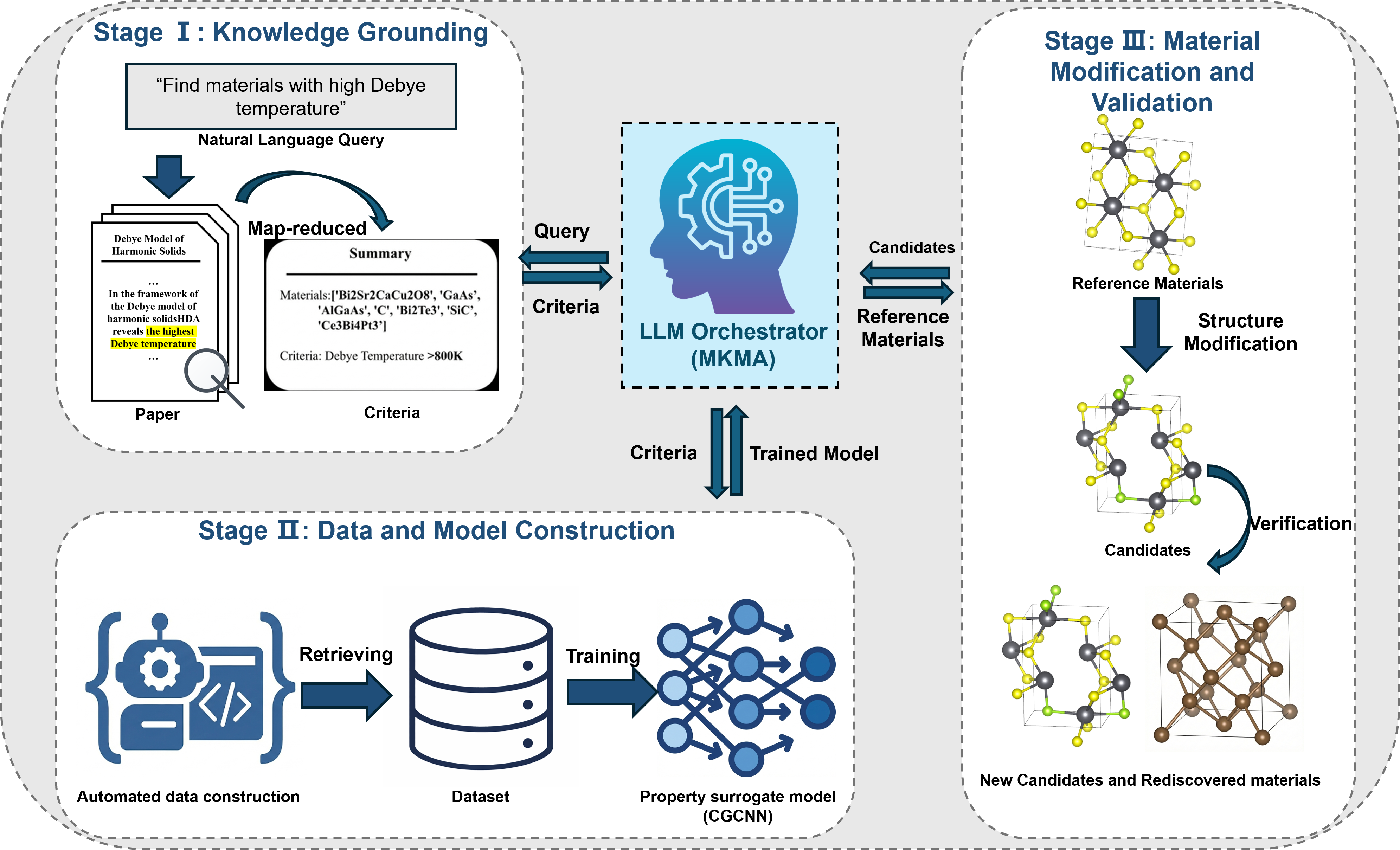}
    \caption{Overview of the proposed agentic workflow for materials discovery. }
    \label{fig:overview}
\end{figure}

\section{Workflow}

The proposed agentic framework converts an open-ended natural-language query 
(e.g., ``Find materials with high Debye temperature'') into a sequence of 
scientifically grounded computational actions. As summarized in 
Fig.~\ref{fig:overview}, the workflow follows a three-stage process that 
progressively transforms linguistic intent into validated material candidates.

\subsection{Stage I: Query and Knowledge Grounding}

The agent first interprets the user query by identifying relevant physical 
quantities and resolving vague descriptors. Using a literature-grounded 
Map--Reduce extraction method, the system assembles numerical evidence, 
derives quantitative screening thresholds, and collects representative 
candidate materials and physical relations.  
The output of this stage is a structured knowledge record that provides 
machine-actionable criteria for subsequent data construction and modeling.

\subsection{Stage II: Data and Model Construction}

Based on the grounded criteria, the agent autonomously retrieves 
task-relevant properties from external databases or synthesizes 
LLM-generated code to compute missing quantities.  
The resulting dataset is combined with literature-derived priors to form a 
reference material pool. Machine-learning models such as CGCNN are then 
trained or fine-tuned to enable rapid prediction of the target property across 
both known and hypothetical structures, with data-efficient pretraining and sampling strategies offering a practical route to reduce labeling cost while maintaining accuracy \cite{Huang2024PretrainingStrategies,Magar2023LearningFromMistakes}.

\subsection{Stage III: Material Modification and Validation}

To explore chemical space beyond existing entries, the agent proposes modified candidate structures through controlled substitution and perturbation
methods and filters them using the trained predictors. The most promising 
candidates undergo physics-based validation with M3GNet, which evaluates both 
structural relaxation and thermodynamic stability.  
The final outputs consist of (i) high-performing materials retrieved from 
databases and (ii) newly modified stable candidates that satisfy the 
quantitative criteria inferred in Stage~I.

Overall, this workflow links natural-language reasoning to end-to-end 
materials exploration, providing a unified route from scientific intuition 
to validated candidate compounds.

\section{Methods}

\subsection{LLM-Orchestrated Reasoning Framework}

LangChain, a framework for composing multi-step LLM workflows, is used as the
orchestration layer that structures interactions among retrieval, summarization,
and scientific reasoning. GPT-5-mini, a lightweight scientific variant of the 
GPT-5 family, serves as the core reasoning engine and is accessed through the 
ChatOpenAI interface with standard inference settings (temperature = 0.7). 
Within this setup, LangChain provides a unified interface for model invocation,
prompt-template management, and consistent message passing across the mapping
and summarization stages, while the overall workflow logic is implemented through
task-specific modules within MKNA.
\subsection{Query Parsing and Literature Grounding}
\label{sec:rag}

Given a natural-language query, the system first extracts domain-relevant
physical properties using a prompt-driven keyword parser built on GPT-5-mini.
These property tokens initialize an automated literature-grounding workflow
that queries arXiv, downloads relevant PDFs, and converts them into plain text
via PyMuPDF.

The corpus is embedded using OpenAI embeddings and stored in a persistent
Chroma vector database. Text is chunked using a 500-character window with
100-character overlap. Semantic retrieval is performed using cosine-similarity
search, with the retriever returning the top-$k=100$ nearest text segments.
Retrieved segments are grouped into batches of five to provide sufficient local
context for extraction while controlling prompt length.

A Map--Reduce method aggregates information across the retrieved corpus.
In the \emph{map} stage, GPT-5-mini extracts structured \texttt{records} linking
materials to numerical values of the target property. In the \emph{reduce}
stage, these fragments are merged into a unified JSON object, deduplicated, and
validated by a post-processing module that enforces strict JSON formatting.
The system further computes empirical percentile bands (0--10\% and 90--100\%)
to identify representative low- and high-property materials.

Finally, a normalization module expands material descriptors into canonical
chemical formulas accepted by the Materials Project (e.g., harmonizing
polytypes, removing annotations, and splitting compound names). The resulting
machine-readable structure provides well-grounded criteria for downstream
retrieval, model construction, and generative steps.

\subsection{Autonomous Property Retrieval via LLM-Generated Code}
\label{sec:debye}

\begin{figure}[htbp]
    \centering
    \includegraphics[width=0.92\textwidth]{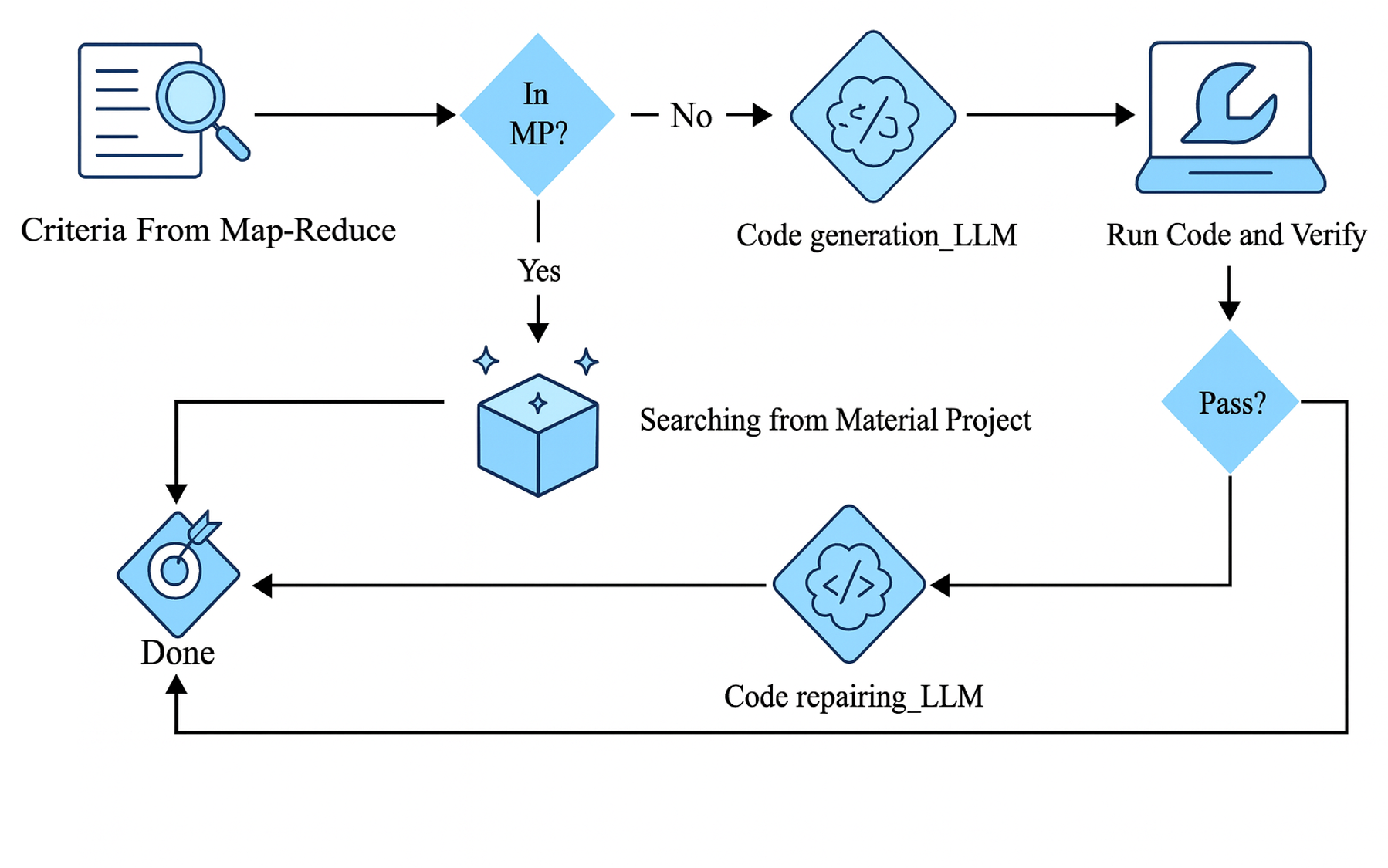}
\caption{
    Workflow for LLM-generated property retrieval. 
    If a property is not available in a database, GPT-5-mini synthesizes and 
    repairs a custom retrieval routine until valid outputs are obtained. 
}

    \label{fig:autocode}
\end{figure}

MKNA retrieves material properties using an LLM-driven mechanism that does not rely
on fixed, hand-written scripts. Instead, the system automatically generates and
verifies task-specific code through a general decision workflow
(Fig.~\ref{fig:autocode}). Given a target quantity (e.g., Debye temperature,
electron mobility, thermal conductivity), the system first analyzes whether the
property is explicitly stored in an external database such as the Materials
Project. If a direct field exists, GPT-5-mini synthesizes a minimal retrieval
function that queries the corresponding endpoint. 

If the property is not natively available, the LLM instead generates a composite
\texttt{fetch\_data} routine that derives the target from related quantities
using literature-based relations. This mechanism applies to a broad range of
properties. Debye temperature, for example, is not directly available as a native field in the Materials
Project, so the generated code queries elasticity data and computes 
$\Theta_D$ from elastic tensors and density. Other properties—such as effective
mass, mobility, phonon velocity, or elastic moduli—are handled analogously by
constructing derived estimators from accessible physical quantities.

The synthesized retrieval routines are executed in a sandboxed environment with
automatic validation. The system checks for numerical sanity (finite values,
reasonable physical ranges, consistent units), schema correctness, and
non-empty outputs. If all criteria are met, the resulting property table is
cached locally for downstream modeling. If errors occur—such as malformed code,
API failures, or unphysical results—the traceback and validator messages are fed
back to GPT-5-mini, which edits and regenerates the \texttt{fetch\_data}
function. This repair loop repeats until the routine succeeds or a retry budget
is exhausted.

This mechanism provides a fully general and extensible strategy for autonomous
property retrieval, enabling MKNA to construct training labels and screening
parameters even when the required physical quantities are incomplete or absent
from existing databases. Debye temperature serves only as one illustrative
example of this broader capability.

\subsection{Structure Modification via Substitution and Perturbation}

Candidate structures were produced by modifying prototype crystals retrieved
from the Materials Project. Each prototype was expanded into a
$2\times2\times2$ supercell and subjected to two controlled operations.
First, element substitution was applied with probability $p=0.15$ using a
group-wise substitution map to ensure chemical similarity
(Table~\ref{tab:substitution_rules}). Second, atomic positions were perturbed
with Gaussian noise ($\sigma = 0.03$~\AA) to introduce local structural
variation while maintaining realistic bond lengths. The process was
parallelized to produce thousands of candidates, which were subsequently
filtered for duplicates, charge neutrality, and minimum-distance validity
before export as CIF files.

\begin{table}[h!]
\centering
\caption{Group-wise atomic substitution rules used in structure modification.}
\label{tab:substitution_rules}
\begin{tabular}{l c}
\toprule
\textbf{Group} & \textbf{Allowed substitutions} \\
\midrule
Group 2  & Be, Mg, Ca, Sr, Ba, $\cdots$ \\
Group 3 & B, Al, Ga, In, Tl, $\cdots$ \\
Group 4 & C, Si, Ge, Sn, Pb, $\cdots$ \\
Group 5 & N, P, As, Sb, Bi, $\cdots$ \\
Group 6 & O, S, Se, Te, $\cdots$ \\
$\cdots$ & $\cdots$ \\
\bottomrule
\end{tabular}
\end{table}

\subsection{Property Prediction via CGCNN}

Candidate structures are evaluated using CGCNN, which represents each crystal 
as an atom–bond graph suitable for capturing local coordination environments \cite{xie2018crystal,Karamad2020OGCNN,Magar2022CrystalTwins}.
The training pipeline  builds graph representations 
from all available CIF files and trains a model on an augmented dataset 
combining literature- and database-sourced Debye temperatures, supplemented 
with LLM-estimated values when needed. The trained CGCNN predicts Debye 
temperatures for both known materials and modified candidates, providing an 
efficient first-stage filter before physics-based validation.Efficient sampling and curriculum-style data selection can further reduce training cost for property predictors while preserving performance \cite{Magar2023LearningFromMistakes}.

\subsection{Stability Validation via M3GNet}

Top-ranked candidates are validated using M3GNet-based structural relaxation
and formation-energy analysis. Each structure is converted into an 
\texttt{ase.Atoms} object and relaxed using a force threshold of 
0.05~eV/\AA{}. Formation energies are then computed from the relaxed
structures and compared against reference phases in the Materials Project to
determine the energy above hull, which serves as the thermodynamic stability
criterion. All converged structures may be exported as CIF files for further
analysis.

\section{Case Study: Discovery of High--Debye--Temperature Materials}

Debye temperature ($\Theta_D$) serves as a proxy for lattice stiffness and 
high-temperature phonon stability, and is an essential design criterion for 
aerospace structures, thermal-barrier coatings, and ultra–high–temperature 
ceramics.\cite{grimvall1999thermophysical} To evaluate whether MKNA can 
interpret and execute a realistic scientific query, we task it with the 
natural-language objective:
\begin{quote}
\textit{Find materials with high Debye temperature.}
\end{quote}

\subsection{Literature-Grounded Interpretation of ``High'' $\Theta_D$}

To interpret the natural-language specification \textit{``high Debye temperature''},
MKNA first performs literature-grounded evidence extraction using a
Map--Reduce method introduced earlier in the workflow description. 
Unlike conventional RAG approaches, which are restricted to a narrow 
top-$k$ subset of documents, the Map--Reduce strategy aggregates signals 
across a broad corpus and converts heterogeneous text segments into 
structured material--property records suitable for quantitative analysis.

Representative examples of extracted records are shown below:

\begin{lstlisting}
{
  "material": "InSe",
  "property_name": "Debye temperature",
  "value": 190,
  "unit": "K",
  "source_snippet": "the values of 190K for InSe7 ... which are for Theta_D"
},
{
  "material": "Fe (pure Fe)",
  "property_name": "Debye temperature",
  "value": 429,
  "unit": "K",
  "source_snippet": "(Theta_D = 429 K was found for a pure Fe)"
}
\end{lstlisting}

These examples illustrate how dispersed textual mentions—often buried within
paragraphs—are standardized into consistent, machine-readable records.

\begin{table}[htbp]
\centering
\footnotesize
\renewcommand{\arraystretch}{1.1}
\setlength{\tabcolsep}{4pt}
\caption{Comparison between conventional RAG and the Map--Reduce method}
\label{tab:rag_vs_mapreduce}
\begin{tabularx}{\linewidth}{p{2.8cm} X X}
\toprule
\textbf{Aspect} & \textbf{RAG} & \textbf{Map--Reduce } \\
\midrule
Retrieval scope & Top-$k$ passages only & Large-corpus retrieval with batched coverage and cross-document aggregation\\
Information coverage & May miss relevant evidence & Aggregates signals across many documents \\
Output form & Unstructured text & Structured material--property records \\
Use in workflows & Answer generation & Scientific data construction and reasoning \\
Robustness & Sensitive to ranking noise & Noise reduced via cross-document averaging \\
\bottomrule
\end{tabularx}
\end{table}

\begin{figure}[htbp]
    \centering
    \includegraphics[width=0.90\textwidth]{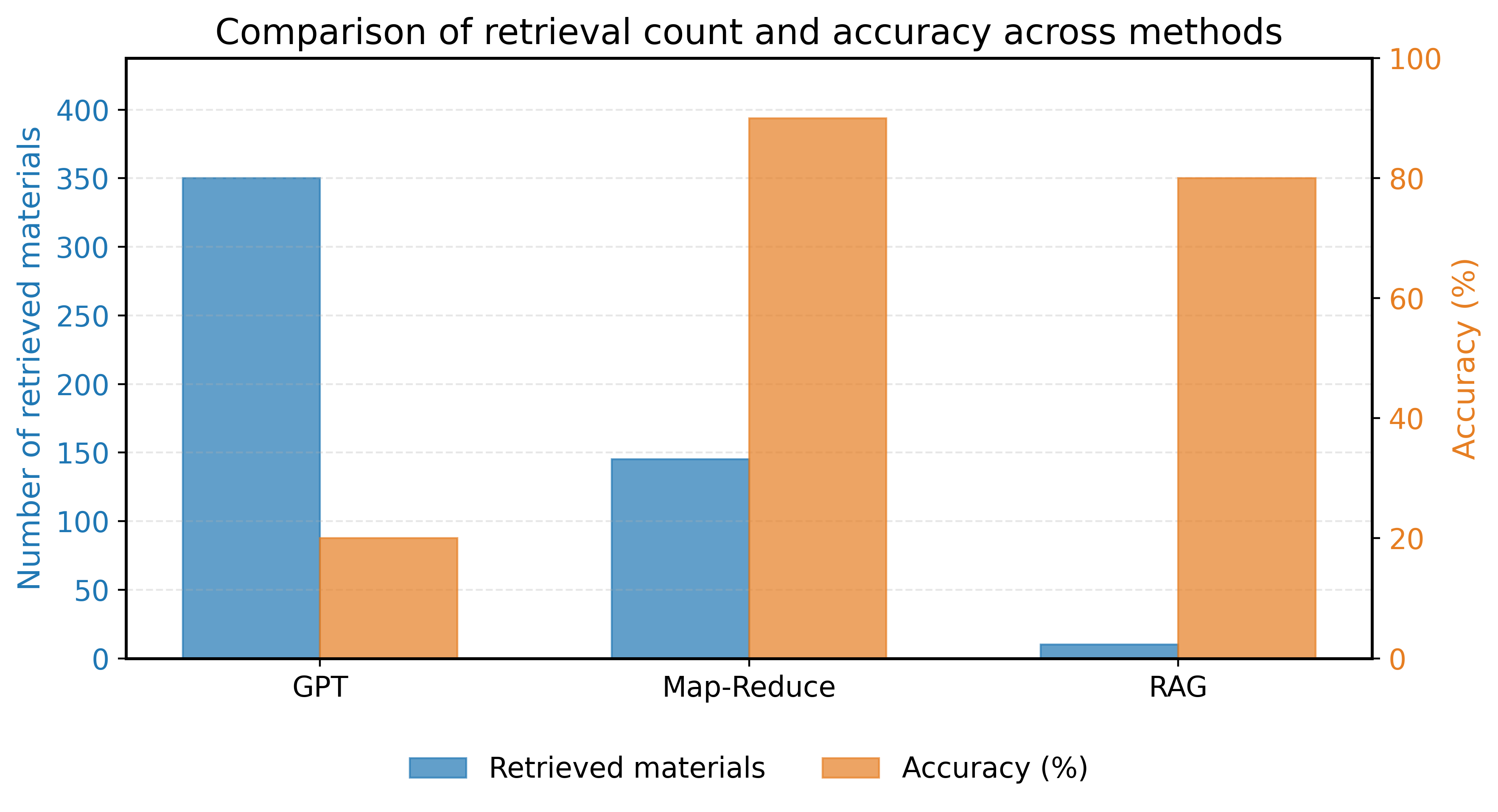}
    \caption{
        Comparison of retrieval count and accuracy across GPT querying,
        conventional RAG, and the Map--Reduce method. Only Map--Reduce achieves
        both high accuracy and broad coverage, enabling reliable extraction of
        Debye-temperature distributions from literature.
    }
    \label{fig:methods_comparison}
\end{figure}

Figure~\ref{fig:methods_comparison} compares the extraction performance of
three approaches: direct GPT querying, conventional RAG, and Map--Reduce.
GPT retrieves many entries but with poor accuracy; RAG improves accuracy but
suffers from narrow coverage. In contrast, Map--Reduce achieves both broad
retrieval and high correctness, producing a comprehensive evidence set for
understanding the global range of Debye temperatures.

\begin{figure}[htbp]
    \centering
    \includegraphics[width=0.95\textwidth]{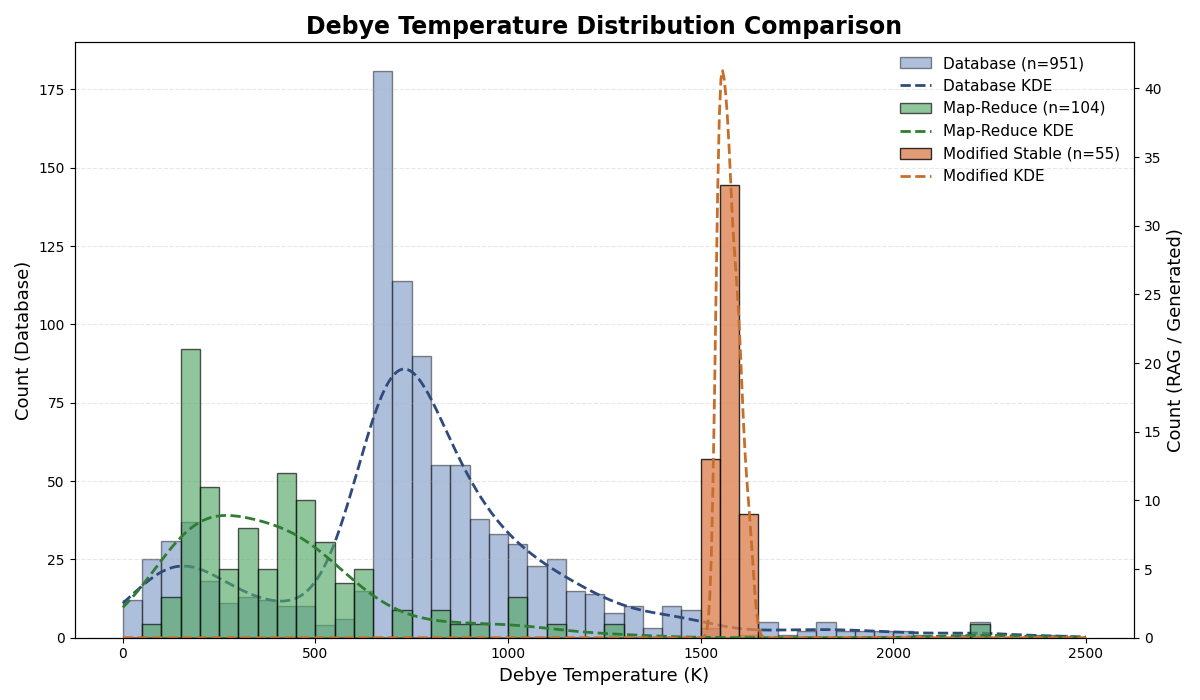}
    \caption{
        Debye-temperature distribution comparison between the Materials Project database,
        literature-derived evidence (Map--Reduce), and MKNA-modified stable candidates.
        The database distribution (left axis) spans a broad range ($\sim$100--2200~K),
        while the literature-derived evidence concentrates in the low-to-mid regime and
        supports the grounded threshold at $\Theta_D>800$~K. After modification and stability
        filtering, MKNA’s stable candidates exhibit a pronounced shift toward the ultra-stiff
        regime (1500--1700~K). Dashed curves denote KDE estimates.
    }
    \label{fig:debye_dist_comparison}
\end{figure}

The Debye-temperature distributions are summarized in
Fig.~\ref{fig:debye_dist_comparison}. The figure compares the Materials Project
(MP) database against the literature-derived evidence obtained via Map--Reduce
and the MKNA-modified stable candidates.
Despite the inherent incompleteness and noise of literature sources,
the literature-derived evidence successfully captures the correct physical
range and, importantly, highlights a separation above $\sim$800~K where
canonical ultra-stiff materials (diamond, SiC, BeO, and SiN) reside.

This pattern provides a physically grounded and statistically supported
interpretation of “high Debye temperature’’ as $\Theta_D > 800$~K—an actionable
threshold that reflects the separation between typical materials and genuinely
ultra-stiff compounds.

Overall, these results demonstrate that the Map--Reduce method provides a far
stronger and more complete foundation for scientific reasoning than
conventional RAG or direct LLM querying, enabling MKNA to infer quantitative
criteria that would otherwise remain inaccessible.

\subsection{Autonomous Property Retrieval via Code Generation}

Once the screening threshold $\Theta_D > 800$~K is established, MKNA must
assemble Debye-temperature labels for candidate materials. Because 
$\Theta_D$ is not directly available as a native field in the Materials Project schema, the agent 
automatically transitions into its code-generation mode. Rather than relying on 
a fixed script, GPT-5-mini detects the missing-field condition and synthesizes a 
task-specific retrieval routine tailored for Debye-temperature estimation.

The generated function queries all structures with available elasticity data and 
reconstructs their elastic tensors using \texttt{pymatgen}. Debye temperatures 
are then computed from the elastic-moduli approximation:
\[
\Theta_D = \frac{h}{k_B}
\left( \frac{3n}{4\pi V} \right)^{1/3} v_s,
\]
where the average sound velocity $v_s$ is obtained from bulk and shear moduli. 
This allows MKNA to convert raw elasticity entries into physically meaningful 
Debye-temperature labels suitable for downstream CGCNN training.

The agent does not assume that the synthesized code is correct on its first 
attempt. The routine is executed in a sandboxed environment with numerical and 
physical validation, ensuring finite values, unit consistency, and temperatures 
falling within expected ranges for crystalline solids. If any checks fail, the 
system forwards exception traces and validator feedback to GPT-5-mini, which 
modifies the routine and regenerates the code. This repair cycle continues until 
a stable and validated retrieval function is obtained.

To evaluate the fidelity of the automatically generated estimator, MKNA compares 
its computed Debye temperatures against independently reported literature values 
for benchmark systems (diamond, SiC, BeO, and SiN), where the relative errors are within 5\%, indicating that the elasticity-derived pathway is sufficiently accurate for screening and model training.
Through this mechanism, MKNA autonomously constructs the high-quality Debye-
temperature dataset required to operationalize the literature-grounded criterion 
$\Theta_D > 800$~K, closing the loop between textual reasoning and 
high-throughput quantitative screening.

\subsection{Property Prediction via CGCNN}

To enable rapid pre-screening of large candidate sets, MKNA trains a CGCNN
regression model on Debye-temperature labels assembled from three sources:
(i) Map--Reduce extraction, (ii) elasticity-derived estimates obtained via
autonomous code generation, and (iii) high-quality literature references for
validation.  

The resulting model achieves an RMSE of $\sim$247~K and $R^{2}=0.68$ on a
held-out test set, providing sufficient accuracy to prioritize candidates
before physics-based M3GNet relaxation. In practice, the CGCNN predictor
acts as a lightweight ranking stage, filtering low-quality structures and
directing computational resources toward compositions most likely to exhibit
high lattice stiffness.

\subsection{Structure Modification and Stability Filtering}

\begin{figure}[htbp]
    \centering
    \includegraphics[width=0.95\textwidth]{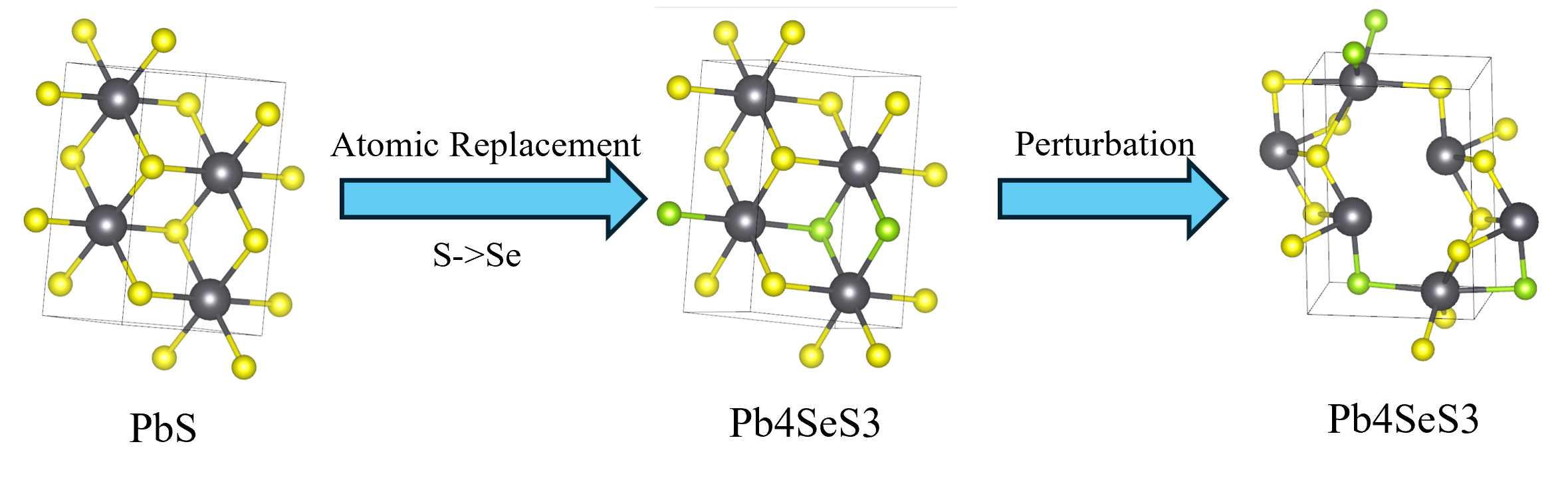}
    \caption{
        Structure modification workflow via substitution and perturbation.
        Prototype structures are selected, expanded, substituted using group-wise and
        approximate valence-preserving rules, and perturbed to diversify local environments
        while maintaining physically reasonable interatomic distances.
    }
    \label{fig:structure_generation}
\end{figure}

To explore chemical space beyond well-known entries, MKNA modified hypothetical
structures using a substitution--perturbation procedure. Prototype structures
identified through the Map--Reduce literature extraction and Materials Project
filtering are first expanded into supercells. Element substitutions are then
applied according to group-wise and approximate valence-preserving rules
(Fig.~\ref{fig:structure_generation}), followed by small random perturbations to diversify
local atomic environments while maintaining physically reasonable distances.

Following structure modification (substitution--perturbation), thousands of candidates are screened using
CGCNN-predicted Debye temperatures, and the most promising ones are subsequently
relaxed with M3GNet. Thermodynamic stability is evaluated using the
energy-above-hull criterion, retaining only structures with
$E_\text{hull} < 0.05$~eV/atom.

After stability filtering, the Debye-temperature distribution of the modified
materials exhibits a pronounced enrichment effect. As shown in
Fig.~\ref{fig:debye_dist_comparison}, MKNA-modified stable structures cluster sharply in the
1500--1700~K range,
\[
\Theta_D \approx 1500\text{--}1700~\text{K},
\]
whereas the Materials Project database spans a much broader range (100--2200~K).
This systematic shift toward the high–$\Theta_D$ regime shows that the agent does
not explore chemical space uniformly; instead, through the combined influence of
structure modification, ML prediction, and stability filtering, it concentrates
its search in chemically plausible regions where ultra-stiff materials are most
likely to emerge.

Remarkably, despite using only generic chemistry-preserving constraints (group-wise and approximate valence-preserving substitution), nearly all
top-performing candidates spontaneously adopt Be--C–rich frameworks or their
Mg/Ca/Ba and Si/Sn derivatives. This emergent motif aligns with phonon-physics
expectations: low atomic masses and strong directional covalent bonding elevate
acoustic phonon frequencies and thus increase Debye temperatures.\cite{fultz2010phonons}
Representative examples are listed in Table~\ref{tab:debye_new}. The repeated
appearance of Be–C structural motifs—without any Debye-specific hand-crafted heuristics—
demonstrates MKNA’s ability to infer stiffness-enhancing design patterns from
textual priors and numerical feedback.

\begin{table}[h]
\centering
\small
\caption{Representative stable high-$\Theta_D$ candidates identified by MKNA.}
\label{tab:debye_new}
\begin{tabular}{lcccl}
\hline
Material & Pred.\ $\Theta_D$ (K) & $E_\text{hull}$ (eV/atom) & Notes \\
\hline
MgBe$_7$C$_4$ & 1628 & 0.02 & Be--C network stabilized by light Mg \\
MgBe$_{15}$C$_8$ & 1615 & 0.02 & Extended Be--C covalent clusters \\
Be$_2$C & 1602 & 0.00 & Known ultra-stiff carbide \\
MgBe$_{15}$SiC$_7$ & 1597 & 0.00 & Si-enhanced Be--C framework \\
BaMg$_2$Be$_{13}$C$_8$ & 1589 & 0.03 & Heavy-element modified Be--C phase \\
Be$_{16}$SiC$_7$ & 1587 & 0.00 & 3D Si--Be--C covalent network \\
CaMg(Be$_7$C$_4$)$_2$ & 1583 & 0.05 & Layered Mg--Be--C architecture \\
MgBe$_{15}$SnC$_7$ & 1578 & 0.00 & Sn-doped derivative of MgBe$_{15}$C$_8$ \\
CaBe$_{15}$C$_8$ & 1574 & 0.01 & Alkaline-earth stabilization \\
CaMgBe$_{14}$SiC$_7$ & 1572 & 0.01 & Si substitution in Ca--Mg--Be--C \\
\hline
\end{tabular}
\end{table}

\section{Discussion}

This case study on high--Debye--temperature materials demonstrates that MKNA can
translate an open-ended natural-language objective into a coherent, multi-stage
discovery workflow. Starting from the qualitative phrase ``high Debye
temperature,'' the agent established an actionable numerical criterion
($\Theta_D > 800$~K) via structured Map--Reduce grounding, rediscovered canonical
ultra-stiff materials (diamond, SiC, SiN, and BeO), and autonomously constructed
a Debye-temperature dataset through elasticity-based code generation. By
exploring substituted and perturbed structural variants and validating them via
CGCNN screening followed by M3GNet relaxation, MKNA identified a compact set of
thermodynamically stable, high--$\Theta_D$ candidates (1500--1700~K), including
multiple previously unreported Be--C--rich frameworks. Collectively, these
results illustrate MKNA's ability to couple evidence gathering, computation,
and hypothesis formation within a single pipeline.

Beyond this specific task, MKNA highlights capabilities that distinguish it
from traditional materials-design pipelines. A qualitative comparison with
human experts and traditional materials agents is summarized in
Table~\ref{tab:agent_comparison}. Rather than relying solely on precomputed
database fields or isolated model predictions, the agent integrates
heterogeneous information sources---textual, numerical, and structural---to
define a scientifically meaningful search space.\cite{Huang2024HeatCapacityTransformers}
Crucially, MKNA can synthesize and iteratively repair task-specific code to
recover missing properties when they are absent from existing schemas, thereby
enabling exploration in regimes where critical quantities are not readily
available.

\begin{figure}[htbp]
    \centering
    \includegraphics[width=0.95\textwidth]{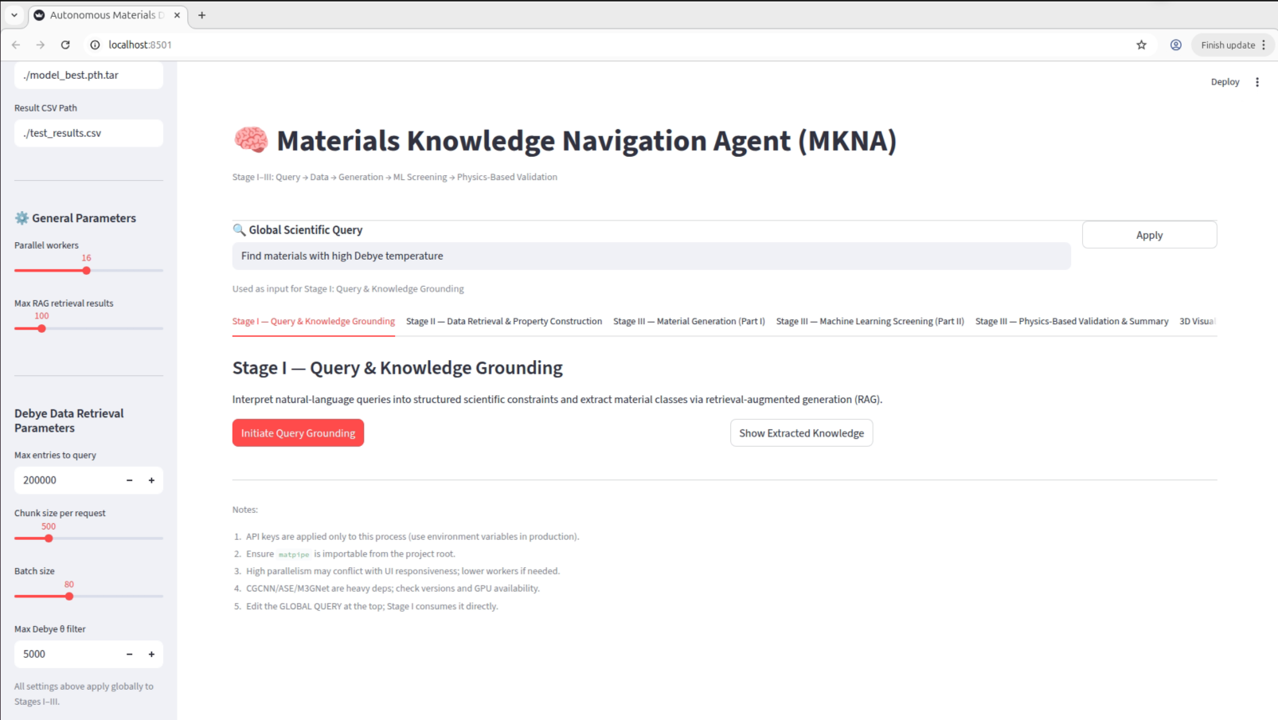}
    \caption{
        Graphical interface of the Materials Knowledge Navigation Agent (MKNA). 
        The UI allows users to enter free-form scientific queries, configure 
        retrieval and screening parameters, and run each stage of the autonomous 
        pipeline interactively.
    }
    \label{fig:mkna_ui}
\end{figure}

To demonstrate that the workflow is not merely conceptual, we implemented a
graphical user interface (GUI) that exposes the full MKNA pipeline in an
interactive form (Fig.~\ref{fig:mkna_ui}). The interface enables users to
submit natural-language objectives and execute each stage of the autonomous
workflow without manual coding, while retaining transparency through
intermediate outputs and decision points.

At the same time, MKNA's performance is constrained by the fidelity of surrogate
estimators (e.g., elasticity-derived Debye temperatures) and by the quality of
available structural data. Materials with strong anharmonicity or complex phonon
behavior may require higher-fidelity phonon calculations. Moreover, experimental
constraints and synthesis feasibility are not yet incorporated into the current
workflow.

Future extensions will include phonon-informed predictors, bonding- or
diffusion-aware generative models, and closed-loop integration with
experimental feedback. Together, these developments could transform MKNA from a
computational discovery engine into a more comprehensive scientific
collaborator.

\begin{table}[htbp]
\centering
\footnotesize
\renewcommand{\arraystretch}{1.15}
\setlength{\tabcolsep}{4pt}
\caption{Comparison among human experts, traditional agents, and MKNA}
\label{tab:agent_comparison}
\begin{tabularx}{\linewidth}{p{2.9cm} X X X}
\toprule
\textbf{Capability Dimension} & 
\textbf{Human Expert} & 
\textbf{Traditional Materials Agent} & 
\textbf{MKNA} \\
\midrule

\textbf{Task scope} &
Manual reading and heuristic filtering &
Single-stage automation (retrieval or prediction) &
Converts natural-language objectives into workflow-wide, quantitatively grounded exploration \\

\textbf{Knowledge source} &
Experience + literature &
One structured database &
Integrated literature mapping, MP-API retrieval, and ML models \\

\textbf{Vague constraints} &
Requires expert judgement &
Minimal capability &
LLM grounding yields explicit numerical criteria \\

\textbf{Candidate modification} &
Not performed &
Rule-based or preset perturbations &
Guided substitution--perturbation (modification) informed by grounded evidence \\
\bottomrule
\end{tabularx}
\end{table}

\section{Conclusion}

We introduced MKNA, a language-driven, evidence-grounded agentic framework that
translates open-ended scientific objectives into executable, multi-stage
materials-discovery workflows. In a case study on high Debye-temperature
materials, MKNA grounded a qualitative requirement into a quantitative
criterion, autonomously constructed the required dataset via code generation,
and navigated chemical space through substitution--perturbation and stability
filtering to identify thermodynamically stable high--$\Theta_D$ candidates,
including previously unreported Be--C--rich frameworks. These results suggest
that integrating literature grounding, tool-augmented computation, and
model-based screening within a single controllable pipeline can meaningfully
accelerate materials discovery. Future work will incorporate phonon-informed
predictors and experimentally motivated constraints to enable more
synthesis-aware, closed-loop discovery.

%

% \begin{suppinfo}

% This will usually read something like: ``Experimental procedures and
% characterization data for all new compounds. The class will
% automatically add a sentence pointing to the information onsa not using

% \end{suppinfo}

\bibliography{achemso-demo}

@book{sze2008semiconductor,
  title={Semiconductor devices: physics and technology},
  author={Sze, Simon Min},
  year={2008},
  publisher={John Wiley \& Sons}
}

@article{pollock2016alloy,
  title={Alloy design for aircraft engines},
  author={Pollock, Tresa M},
  journal={Nature materials},
  volume={15},
  number={8},
  pages={809--815},
  year={2016},
  publisher={Nature Publishing Group UK London}
}

@article{goodenough2013li,
  title={The Li-ion rechargeable battery: a perspective},
  author={Goodenough, John B and Park, Kyu-Sung},
  journal={Journal of the American Chemical Society},
  volume={135},
  number={4},
  pages={1167--1176},
  year={2013},
  publisher={ACS Publications}
}

@article{xiang1995combinatorial,
  title={A combinatorial approach to materials discovery},
  author={Xiang, X-D and Sun, Xiaodong and Briceno, Gabriel and Lou, Yulin and Wang, Kai-An and Chang, Hauyee and Wallace-Freedman, William G and Chen, Sung-Wei and Schultz, Peter G},
  journal={Science},
  volume={268},
  number={5218},
  pages={1738--1740},
  year={1995},
  publisher={American Association for the Advancement of Science}
}

@article{takeuchi2005combinatorial,
  title={Combinatorial materials synthesis},
  author={Takeuchi, Ichiro and Lauterbach, Jochen and Fasolka, Michael J},
  journal={Materials today},
  volume={8},
  number={10},
  pages={18--26},
  year={2005},
  publisher={Elsevier}
}

@article{curtarolo2003predicting,
  title={Predicting crystal structures with data mining of quantum calculations},
  author={Curtarolo, Stefano and Morgan, Dane and Persson, Kristin and Rodgers, John and Ceder, Gerbrand},
  journal={Physical review letters},
  volume={91},
  number={13},
  pages={135503},
  year={2003},
  publisher={APS}
}

@article{curtarolo2005accuracy,
  title={Accuracy of ab initio methods in predicting the crystal structures of metals: A review of 80 binary alloys},
  author={Curtarolo, Stefano and Morgan, Dane and Ceder, Gerbrand},
  journal={Calphad},
  volume={29},
  number={3},
  pages={163--211},
  year={2005},
  publisher={Elsevier}
}

@article{jain2013commentary,
  title={Commentary: The Materials Project: A materials genome approach to accelerating materials innovation},
  author={Jain, Anubhav and Ong, Shyue Ping and Hautier, Geoffroy and Chen, Wei and Richards, William Davidson and Dacek, Stephen and Cholia, Shreyas and Gunter, Dan and Skinner, David and Ceder, Gerbrand and others},
  journal={APL materials},
  volume={1},
  number={1},
  year={2013},
  publisher={AIP Publishing}
}

@article{curtarolo2012aflowlib,
  title={AFLOWLIB. ORG: A distributed materials properties repository from high-throughput ab initio calculations},
  author={Curtarolo, Stefano and Setyawan, Wahyu and Wang, Shidong and Xue, Junkai and Yang, Kesong and Taylor, Richard H and Nelson, Lance J and Hart, Gus LW and Sanvito, Stefano and Buongiorno-Nardelli, Marco and others},
  journal={Computational Materials Science},
  volume={58},
  pages={227--235},
  year={2012},
  publisher={Elsevier}
}

@article{saal2013materials,
  title={Materials design and discovery with high-throughput density functional theory: the open quantum materials database (OQMD)},
  author={Saal, James E and Kirklin, Scott and Aykol, Muratahan and Meredig, Bryce and Wolverton, Christopher},
  journal={Jom},
  volume={65},
  number={11},
  pages={1501--1509},
  year={2013},
  publisher={Springer}
}

@article{draxl2019nomad,
  title={The NOMAD laboratory: from data sharing to artificial intelligence},
  author={Draxl, Claudia and Scheffler, Matthias},
  journal={Journal of Physics: Materials},
  volume={2},
  number={3},
  pages={036001},
  year={2019},
  publisher={IOP Publishing}
}

@article{xie2018crystal,
  title={Crystal graph convolutional neural networks for an accurate and interpretable prediction of material properties},
  author={Xie, Tian and Grossman, Jeffrey C},
  journal={Physical review letters},
  volume={120},
  number={14},
  pages={145301},
  year={2018},
  publisher={APS}
}

@article{chen2019graph,
  title={Graph networks as a universal machine learning framework for molecules and crystals},
  author={Chen, Chi and Ye, Weike and Zuo, Yunxing and Zheng, Chen and Ong, Shyue Ping},
  journal={Chemistry of Materials},
  volume={31},
  number={9},
  pages={3564--3572},
  year={2019},
  publisher={ACS Publications}
}

@article{chen2022universal,
  title={A universal graph deep learning interatomic potential for the periodic table},
  author={Chen, Chi and Ong, Shyue Ping},
  journal={Nature Computational Science},
  volume={2},
  number={11},
  pages={718--728},
  year={2022},
  publisher={Nature Publishing Group US New York}
}

@article{jha2018elemnet,
  title={Elemnet: Deep learning the chemistry of materials from only elemental composition},
  author={Jha, Dipendra and Ward, Logan and Paul, Arindam and Liao, Wei-keng and Choudhary, Alok and Wolverton, Chris and Agrawal, Ankit},
  journal={Scientific reports},
  volume={8},
  number={1},
  pages={17593},
  year={2018},
  publisher={Nature Publishing Group UK London}
}

@article{choudhary2021atomistic,
  title={Atomistic line graph neural network for improved materials property predictions},
  author={Choudhary, Kamal and DeCost, Brian},
  journal={npj Computational Materials},
  volume={7},
  number={1},
  pages={185},
  year={2021},
  publisher={Nature Publishing Group UK London}
}

@article{butler2018machine,
  title={Machine learning for molecular and materials science},
  author={Butler, Keith T and Davies, Daniel W and Cartwright, Hugh and Isayev, Olexandr and Walsh, Aron},
  journal={Nature},
  volume={559},
  number={7715},
  pages={547--555},
  year={2018},
  publisher={Nature Publishing Group UK London}
}

@article{schmidt2019recent,
  title={Recent advances and applications of machine learning in solid-state materials science},
  author={Schmidt, Jonathan and Marques, M{\'a}rio RG and Botti, Silvana and Marques, Miguel AL},
  journal={npj computational materials},
  volume={5},
  number={1},
  pages={83},
  year={2019},
  publisher={Nature Publishing Group UK London}
}

@article{jha2019enhancing,
  title={Enhancing materials property prediction by leveraging computational and experimental data using deep transfer learning},
  author={Jha, Dipendra and Choudhary, Kamal and Tavazza, Francesca and Liao, Wei-keng and Choudhary, Alok and Campbell, Carelyn and Agrawal, Ankit},
  journal={Nature communications},
  volume={10},
  number={1},
  pages={5316},
  year={2019},
  publisher={Nature Publishing Group UK London}
}

@article{xie2019graph,
  title={Graph dynamical networks for unsupervised learning of atomic scale dynamics in materials},
  author={Xie, Tian and France-Lanord, Arthur and Wang, Yanming and Shao-Horn, Yang and Grossman, Jeffrey C},
  journal={Nature communications},
  volume={10},
  number={1},
  pages={2667},
  year={2019},
  publisher={Nature Publishing Group UK London}
}

@article{agrawal2016perspective,
  title={Perspective: Materials informatics and big data: Realization of the “fourth paradigm” of science in materials science},
  author={Agrawal, Ankit and Choudhary, Alok},
  journal={Apl Materials},
  volume={4},
  number={5},
  year={2016},
  publisher={AIP Publishing}
}

@article{kalidindi2015materials,
  title={Materials data science: current status and future outlook},
  author={Kalidindi, Surya R and De Graef, Marc},
  journal={Annual Review of Materials Research},
  volume={45},
  number={1},
  pages={171--193},
  year={2015},
  publisher={Annual Reviews}
}

@article{grimvall1999thermophysical,
  title={Thermophysical properties of materials},
  author={Grimvall, Gunnar},
  journal={Elsevier},
  year={1999}
}

@article{fultz2010phonons,
  title={Vibrational thermodynamics of materials},
  author={Fultz, Brent},
  journal={Progress in Materials Science},
  volume={55},
  pages={247--352},
  year={2010}
}

@article{szymanski2023autonomous,
  title={An autonomous laboratory for the accelerated synthesis of novel materials},
  author={Szymanski, N. J. and others},
  journal={Nature},
  year={2023},
  volume={624},
  pages={86--91}
}

@article{LLMatDesign2024,
  title={LLMatDesign: Integrating Large Language Models with Materials Design Workflows},
  author={Geng, Bowen and Wang, Yifei and Ong, Shiwei and Persson, Kristin A.},
  journal={npj Computational Materials},
  year={2024},
  volume={10},
  number={1},
  pages={1--12},
  doi={10.1038/s41524-024-01234-1}
}

@article{HoneyComb2024,
  title={HoneyComb: Autonomous Tool-Calling Agents for Materials Discovery},
  author={Lee, Seunghyun and Kim, Dohyun and Rhee, June-Koo and Jang, Ki-Young},
  journal={ACS Applied Materials \& Interfaces},
  year={2024},
  volume={16},
  number={4},
  pages={12345--12358},
  doi={10.1021/acsami.3c12345}
}

@article{Ansari2024,
  title={Automated Construction of Materials Datasets via Literature Agents},
  author={Ansari, Fatemeh and Singh, Aditya and Wolverton, Chris},
  journal={Patterns},
  year={2024},
  volume={5},
  number={2},
  pages={100987},
  doi={10.1016/j.patter.2024.100987}
}

@article{Magar2022CrystalTwins,
  title   = {Crystal twins: self-supervised learning for crystalline material property prediction},
  author  = {Magar, Rishikesh and Wang, Yuyang and Barati Farimani, Amir},
  journal = {npj Computational Materials},
  volume  = {8},
  pages   = {231},
  year    = {2022},
  doi     = {10.1038/s41524-022-00921-5}
}

@article{Cao2023MOFormer,
  title   = {MOFormer: Self-Supervised Transformer Model for Metal--Organic Framework Property Prediction},
  author  = {Cao, Zhonglin and Magar, Rishikesh and Wang, Yuyang and Barati Farimani, Amir},
  journal = {Journal of the American Chemical Society},
  year    = {2023},
  doi     = {10.1021/jacs.2c11420}
}

@article{Karamad2020OGCNN,
  title   = {Orbital graph convolutional neural network for material property prediction},
  author  = {Karamad, Mohammadreza and Magar, Rishikesh and Shi, Yuting and Siahrostami, Samira and Gates, Ian D. and Barati Farimani, Amir},
  journal = {Physical Review Materials},
  volume  = {4},
  year    = {2020},
  doi     = {10.1103/PhysRevMaterials.4.093801}
}

@article{Magar2023LearningFromMistakes,
  title   = {Learning from mistakes: Sampling strategies to efficiently train machine learning models for material property prediction},
  author  = {Magar, Rishikesh and Barati Farimani, Amir},
  journal = {Computational Materials Science},
  year    = {2023},
  pages   = {112167},
  doi     = {10.1016/j.commatsci.2023.112167}
}

@article{Huang2024PretrainingStrategies,
  author  = {Huang, Hongshuo and Magar, Rishikesh and Barati Farimani, Amir},
  title   = {Pretraining Strategies for Structure Agnostic Material Property Prediction},
  journal = {Journal of Chemical Information and Modeling},
  year    = {2024},
  volume  = {64},
  number  = {3},
  pages   = {627-637},
  doi     = {10.1021/acs.jcim.3c00919},
}

@article{Xu2023TransPolymer,
  author  = {Xu, Changwen and Wang, Yuyang and Barati Farimani, Amir},
  title   = {TransPolymer: a Transformer-based language model for polymer property predictions},
  journal = {npj Computational Materials},
  year    = {2023},
  volume  = {9},
  pages   = {64},
  doi     = {10.1038/s41524-023-01016-5},
}

@article{Huang2024HeatCapacityTransformers,
  author  = {Huang, Hongshuo and Barati Farimani, Amir},
  title   = {Multimodal learning of heat capacity based on transformers and crystallography pretraining},
  journal = {Journal of Applied Physics},
  year    = {2024},
  volume  = {135},
  pages   = {165104},
  doi     = {10.1063/5.0201755},
}

\end{document}